\documentclass[letterpaper]{article} 
\usepackage{aaai25}  
\usepackage{times}  
\usepackage{helvet}  
\usepackage{courier}  
\usepackage[hyphens]{url}  
\usepackage{graphicx} 
\usepackage{natbib}  
\usepackage{caption} 
\usepackage{algorithm}
\usepackage{algorithmic}
\usepackage{subfigure}
\usepackage{multirow}
\usepackage{makecell}
\usepackage{dcolumn}
\usepackage{booktabs}
\usepackage{amsmath}
\usepackage{amsfonts}

\usepackage{newfloat}
\usepackage{listings}
\DeclareCaptionStyle{ruled}{labelfont=normalfont,labelsep=colon,strut=off} 
\floatstyle{ruled}
\newfloat{listing}{tb}{lst}{}
\floatname{listing}{Listing}
\title{SkillTree: Explainable Skill-Based Deep Reinforcement Learning \\for Long-Horizon Control Tasks}
\author{
    Yongyan Wen\textsuperscript{\rm 1},
    Siyuan Li\textsuperscript{\rm 1}\footnote{Corresponding author.},
    Rongchang Zuo\textsuperscript{\rm 1},
    Lei Yuan\textsuperscript{\rm 2, 3, 4},
    Hangyu Mao\textsuperscript{\rm 5},
    Peng Liu\textsuperscript{\rm 1}
}
\affiliations{
    \textsuperscript{\rm 1}Faculty of Computing, Harbin Institute of Technology\\
    \textsuperscript{\rm 2}National Key Laboratory of Novel Software Technology, Nanjing University\\
    \textsuperscript{\rm 3}School of Artificial Intelligence, Nanjing University\\
    \textsuperscript{\rm 4}Polixir Technologies,
    \textsuperscript{\rm 5}SenseTime Research\\
    23B903027@stu.hit.edu.cn, siyuanli@hit.edu.cn, 2021110788@stu.hit.edu.cn, yuanl@lamda.nju.edu.cn, maohangyu@sensetime.com, pengliu@hit.edu.cn
}

\usepackage{bibentry}

\begin{document}

\maketitle

\begin{abstract}
    Deep reinforcement learning (DRL) has achieved remarkable success in various domains, yet its reliance on neural networks results in a lack of transparency, which limits its practical applications in safety-critical and human-agent interaction domains. Decision trees, known for their notable explainability, have emerged as a promising alternative to neural networks. However, decision trees often struggle in long-horizon continuous control tasks with high-dimensional observation space due to their limited expressiveness. To address this challenge, we propose SkillTree, a novel hierarchical framework that reduces the complex continuous action space of challenging control tasks into discrete skill space. By integrating the differentiable decision tree within the high-level policy, SkillTree generates diecrete skill embeddings that guide low-level policy execution. Furthermore, through distillation, we obtain a simplified decision tree model that improves performance while further reducing complexity. Experiment results validate SkillTree's effectiveness across various robotic manipulation tasks, providing clear skill-level insights into the decision-making process. The proposed approach not only achieves performance comparable to neural network based methods in complex long-horizon control tasks but also significantly enhances the transparency and explainability of the decision-making process.
\end{abstract}

%

\section{Introduction}
Deep Reinforcement learning (DRL) has been shown as a powerful framework for tackling complex decision-making tasks, achieving remarkable success in various domains such as games \cite{mnih2015human, baai2023plan4mc}, robotic manipulation \cite{akkaya2019solving, jitosho2023reinforcement}, and visual navigation \cite{kulhanek2021visual}. Despite these advancements, the black-box nature of neural networks poses significant challenges in understanding and trusting their decision making processes. This lack of transparency is particularly concerning in safety-sensitive and human-agent interaction applications, where understanding the rationale behind decisions is essential \cite{hickling2023explainability}. For example, in the autonomous driving domain, only if human users can understand the driving policies can they trustfully use them in their cars.


Explainable reinforcement learning (XRL) \cite{heuillet2021explainability, hickling2023explainability, milani2024explainable} aims to enhance the transparency and explainability of DRL models. XRL methods can be broadly categorized into inherently explainable models, where explainability is induced in the model training process, and post-hoc explanations, where models are explained after training. Existing works have shown that using a decision tree (DT) \cite{costa2023recent} as the underlying model is effective \cite{frosst2017distilling, bastani2018verifiable, ding2020cdt, vasicETAL22MoET}. 
In a DT, observations are input at the root node, and decisions are determined by conditional branches leading to the leaf nodes, which provide the final outputs. This simple structure gives DTs high explainability, offering clear and straightforward decision paths. Unlike post-hoc methods that rely on external algorithms to interpret black-box models, DT-based methods embed transparency directly within the model. The DT-based XRL paradigm not only enhances clarity but also simplifies debugging and validation, as the rationale behind each decision is explicitly encoded within the tree.

Despite their advantages, DT-based methods are not suitable for the following challenging task settings. (a) \textbf{Long-horizon tasks}: The temporally-extended decision processes require large and complex trees, which are difficult to optimize \cite{liu2021learning}. (b) \textbf{High-dimensional state spaces}: DTs often lack sufficient representational ability to effectively manage high-dimensional state spaces, leading to suboptimal performance in these environments \cite{bastani2018verifiable, ding2020cdt}. (c) \textbf{Continuous action spaces}: The limited number of leaf nodes constrains DTs' ability to encode continuous control policies optimally, particularly in complex robotic tasks. These limitations restrict the applicability of DT-based RL methods in complex environments, highlighting the need for approaches that can manage high-dimensional spaces and long-horizon tasks.

To address the limitations of traditional DTs in handling long-horizon, high-dimensional, and continuous control tasks, we propose a novel framework called SkillTree. SkillTree introduces the concept of skills, which represent temporal abstractions of low-level actions in the context of RL. These skills simplify the decision-making process by breaking down long trajectories into manageable segments, enabling the agent to plan at a higher level, thereby reducing the complexity associated with extended tasks. However, the challenge of skill representation arises, as skills are often represented in continuous spaces, leading to difficulties in control and explainability. To overcome this, our framework leverages hierarchical structures and discrete skill representation learning. By regularizing the skill space into discrete units, we simplify policy learning for skill selection and enhance the explainability of the learned skills. Specifically, we employ decision trees to index and execute these discrete skills, combining the inherent explainability of decision trees with the flexibility of skill-based representations. This approach provides a robust solution for managing complex, long-horizon decision tasks while offering clear, skill-level insights into the decision-making process. We summarize the main contributions of this paper as follows:
\begin{enumerate}
    \item We propose SkillTree, a novel hierarchical, skill-based method for explainable reinforcement learning. To the best of our knowledge, it marks the first successful application of DT-based explainable method in long-horizon continuous control tasks.
    \item We introduce a method for discrete skill representation learning, which effectively reduces the skill space and improves the efficiency of skill-based policy learning.
    \item Experiment results across a variety of robotic manipulation tasks demonstrate that our method provides skill-level explanations while achieving performance comparable to neural network based approaches.
\end{enumerate}

\section{Related Work}
\subsection{Explainable RL}
Recently, explainability approaches in DRL have been broadly categorized into intrinsic and post-hoc explanations, based on the method of their generation. We briefly discuss these two classes of methods here. For a more detailed taxonomy and discussion of XRL, please refer to the following surveys: \cite{heuillet2021explainability, hickling2023explainability, milani2024explainable}.

Intrinsic explanation methods are designed with explainability as a fundamental aspect, often incorporating simpler, more transparent models such as decision trees \cite{liu2019toward, silva2020optimization, ding2020cdt}, linear models \cite{rajeswaran2017towards, molnar2020interpretable, wabartha2023piecewise} or symbolic rules \cite{lyu2019sdrl, ma2021learning}. By ensuring that the model itself is explainable, intrinsic methods offer real-time, straightforward explanations of the agent’s decisions. For instance, using a decision tree as the policy model enables immediate and clear understanding of the decision-making process.

Post-hoc methods apply explainability techniques after the RL model has been trained. These approaches do not alter the original model but instead seek to explain the decisions of more complex, often black-box models. Typical post-hoc methods include feature attribution methods like saliency maps \cite{greydanus2018visualizing, anderson2019explaining, olson2021counterfactual} and Shapley value \cite{zhang2021explainable, heuillet2022collective}, which highlight the most influential input features. Model distillation approximates the black-box model, either locally or globally, with explainable models like decision trees \cite{bastani2018verifiable, coppens2019distilling, bewley2021tripletree, orfanos2023synthesizing}. Counterfactual explanations generate alternative scenarios by modifying input features to observe changes in the model’s output, offering a way to understand the sensitivity of decisions \cite{madumal2020explainable, olson2021counterfactual}. Some studies have explored identifying critical states crucial for achieving the final reward from a state-reward perspective \cite{guo2021edge, cheng2024statemask}. Additionally, researchers have analyzed the impact of different training samples on policy training outcomes \cite{deshmukh2023explaining}. Lastly, rule extraction methods create human-readable rules that approximate the model’s behavior and enhance the transparency of the decision-making process \cite{hein2018interpretable, landajuela2021discovering}.

\subsection{Decision Tree}
Decision trees are widely used for their explainability and simplicity, often serving as function approximators in reinforcement learning. Classical decision tree algorithms like CART \cite{loh2011classification} and C4.5 \cite{quinlan1993c4} produce explainable surrogate policies but are limited in expressiveness and impractical for integration into DRL models. Approaches such as VIPER \cite{bastani2018verifiable} attempt to distill neural network policies into verifiable decision tree by imitation learning. Recent advancements include rule-based node divisions \cite{dhebar2020interpretable} and parametric differentiable decision tree, such as soft decision tree \cite{frosst2017distilling} and differentiable decision tree for approximating Q-function or policy \cite{silva2020optimization}. While these methods improve expressiveness, they are generally constrained to simpler, low-dimensional environments. By contrast, our approach addresses the complexity of high-dimensional continuous control tasks by converting the action space into a skill space, thereby reducing complexity and providing a more effective and explainable solution for challenging tasks.

\subsection{Skills in Reinforcement Learning}
Recent works have increasingly focused on learning and utilizing skills to improve agent efficiency and generalization \cite{shu2018hierarchical, hausman2018learning, shankar2020learning, lynch2020learning}. Skills can be extracted from the offline dataset \cite{shiarlis2018taco, kipf2019compile, pertsch2021accelerating, pertsch2021demonstration, shi2023skill} or manually defined \cite{lee2019learning, dalal2021accelerating, baai2023plan4mc}. Typically, these skills are represented as high-dimensional, continuous latent variables, which limits their explainability. In contrast, we propose extracting a discrete skill representation from a task-agnostic dataset, making it suitable for DT-based policy learning.

\section{Preliminary}
\subsection{Reinforcement Learning}
In this paper, we are concerned with a finite Markov decision process \cite{sutton2018reinforcement}, which can be represented as a tuple $M=(\mathcal{S},\mathcal{A},R,\mathcal{P},p_0,\gamma,H)$. Where $\mathcal{S}$ is the state space, $\mathcal{A}$ is the action space, $\mathcal{S}\times\mathcal{A}\mapsto\mathbb{R}$ is the reward space, $\mathcal{P}:\mathcal{S}\times\mathcal{A}\mapsto\mathrm{\Delta}(\mathcal{S})$ is the transfer function, $p_0:\Delta(\mathcal{S})$ is the initial state distribution, $\gamma\in(0,1)$ is the discount factor, and $H$ is the episode length. The learning objective is to find the optimal policy $\pi$ maximizing the expected discounted return
\begin{equation}
    \max{\mathcal{J}}(\pi) = \max_\pi \mathbb{E}_{s_0\sim p_0, (s_0,a_0,\dots,s_H)\sim\pi}\left[\sum_{t=0}^{H-1} \gamma^t\mathcal{R}(s_t,a_t)\right].
\end{equation}

\subsection{Soft Decision Tree}
\begin{figure}[t]
    \centering
    \includegraphics[width=\linewidth]{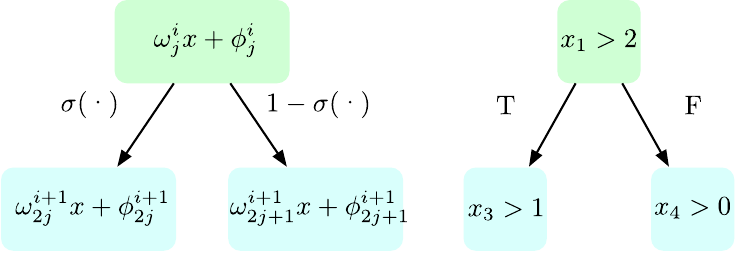}
    \caption{Comparison of the soft decision tree (left) and the hard decision tree (right).}
    \label{fig.decision_tree}
\end{figure}

The differentiable decision tree \cite{frosst2017distilling} is a special type of decision tree that differs from the traditional ones like CART \cite{loh2011classification} by using probabilistic decision boundaries instead of deterministic boundaries. This modification increases the flexibility of model by allowing for smoother transitions at each decision node, as shown in Figure \ref{fig.decision_tree}. The differentiable soft decision tree is a complete binary tree in which each inner node can be represented by a learnable weight $\omega_j^i$ and bias $\phi_j^i$. Here, $i$ and $j$ represent the layer index and the node index of that layer, respectively. By using the sigmoid function $\sigma(\cdot)$, the probability $\sigma(\omega x+\phi)$ represents the transfer probability from that node to the left subtree, and $1-\sigma(\omega x+\phi)$ gives the transfer probability to the right subtree. This process allows the decision tree to provide a ``soft'' decision at each node for the current input.

Consider a decision tree with depth $d$. The nodes in the tree are denoted by $n_u$, where $u=2^i-1+j$ denotes the node index, and the decision path $P$ can be represented as
\begin{equation}\label{eq.decision}
    P= \mathop{\arg\max}\limits_{\{u\}} \prod_{i=0}^{d-1}\prod_{j=0}^{2^i} p_{\left\lfloor\frac{j}{2}\right\rfloor\rightarrow j}^{i\rightarrow i+1}
\end{equation}
where $\{u\}$ denotes nodes on the decision path, and $p_{\lfloor\frac{j}{2}\rfloor\rightarrow j}^{i\rightarrow i+1}$ denotes the probability of moving from node $n_{2^i+\lfloor j/2\rfloor}$ to node $n_{2^{i+1}+j}$, the probability is calculated as
\begin{equation}\label{eq.prob}
    p_{\left\lfloor\frac{j}{2}\right\rfloor \rightarrow j}^{i \rightarrow i+1}=
    \begin{cases}
        \sigma\left(\omega_{\left\lfloor\frac{j}{2}\right\rfloor}^i x+\phi_{\left\lfloor\frac{j}{2}\right\rfloor}^i\right) & \text{if}\ j\mod 2=0, \\
        1-\sigma\left(\omega_{\left\lfloor\frac{j}{2}\right\rfloor}^i x+\phi_{\left\lfloor\frac{j}{2}\right\rfloor}^i\right) & \text{otherwise}.
    \end{cases}
\end{equation}

With Equation \ref{eq.decision} and \ref{eq.prob}, the tree can be traversed down to the leaf nodes. Leaf nodes are represented by parameter vectors $\mathbf{w}^{2^d \times K}$, where $K$ is the number of output categories. The output of the leaf node $k$ is a categorical distribution given by $\operatorname{softmax}(\mathbf{w}_k)$, which is independent of the input. \textit{The learnable parameters of both the leaf nodes and the inner nodes can be optimized using existing DRL algorithms, making the soft decision tree suitable as an explainable alternative model for DRL policy.}

\section{Explainable RL with SkillTree}
Our goal is to leverage discrete skill representations to learn a skill decision tree, enabling skill-level explainability. Overall, our approach is divided into three stages: (1) extracting the discrete skill embeddings from the offline dataset, (2) training an explainable DT-based skill policy by RL for downstream long-horizon tasks, and (3) distilling the trained policy into a simplified decision tree.


\begin{figure*}[t]
    \centering
    \includegraphics[width=0.85\linewidth]{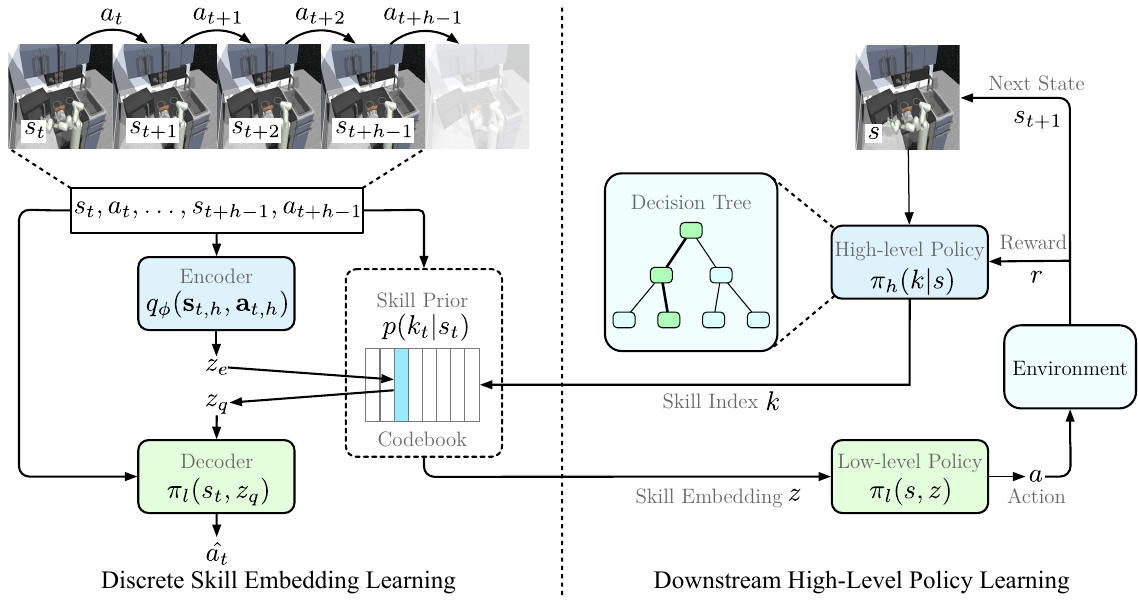}
    \caption{Discrete skill embedding learning and downstream high-level DT policy learning. After completing the skill learning, we freeze the decoder and skill prior, and then proceed to finetune the codebook during the high-level policy learning.}
    \label{fig.framework}
\end{figure*}

\subsection{Learning Discrete Skill Embeddings}
Our goal is to obtain a fixed number of skill embeddings that contain a temporal abstraction of a sequence of actions. Skills represent action sequences of useful behaviors and are represented by $D$-dimensional vectors. We assume an existing task-agnostic dataset $\mathcal{D}=\{\tau_i\}, \tau_i=\{s_0,a_0,\dots,s_{H_i},a_{H_i}\}$, consists of $d$ trajectories of varying lengths, where each trajectory includes states $s_t,\dots,s_{H_i}$ and corresponding actions $a_t,\dots,a_{H_i}$. To regularize the skill embeddings, instead of learning within the continuous skill space, we employ a skill table $Z=\{e_1,e_2,\dots,e_K\}$ (i.e., codebook) that contains $D$-dimensional vectors, each representing a learnable skill embedding.

To learn reusable skills from offline datasets and guide exploration efficiently, we modified VQ-VAE \cite{van2017neural} to learn skill representations and skill prior, as illustrated in Figure \ref{fig.framework}. First, a demonstration trajectory is randomly divided into state-action segments of fixed length $h$ during the training. Each segment contains $h$ observation states $\mathbf{s}_{t,h}=s_t,s_{t+1},\dots,s_{t+h-1}$ and the corresponding actions $\mathbf{a}_{t,h}=a_t,a_{t+1},\dots,a_{t+h-1}$. The input to the encoder $q_\phi$ is $\mathbf{s}_{t,h}$ and $\mathbf{a}_{t,h}$. The output of the encoder is a $D$-dimensional embedding $z_e$. Next, we select the embedding from the codebook that is nearest to $z_e$ and its index $k=\mathop{\arg\min}\limits_{j}\Vert z_e-e_j\Vert_2$, obtaining the regularized embedding $z_q=e_k$. Finally, the low-level actual action $a_t$ is obtained through the state-conditioned decoder $\pi_l(s_t,z_q)$, which serves as the low-level policy.

In addition, following the approach of \cite{pertsch2021accelerating}, we introduce a skill prior to guide the high-level policy. The skill prior $p(k_t|s_t)$ predicts the skill distribution given the first state of the sequence in the offline data. This enhances exploration efficiency during high-level policy training by avoiding random skill sampling during warming up. Here, $s_t$ represents the first state of the sequence. The skill prior is a $K$-categorical distribution that matches the output of encoder. The objective of the skill prior is to fit the output distribution of encoder conditioned solely on the first state $s_t$. Since the encoder can only produce a fixed skill index $k$, the skill prior aims to match a one-hot distribution of that index, aligning its predictions with the encoder. The overall training objective is
\begin{align}
    \mathcal{L} = & \sum_{t=0}^{h-1}\Vert\pi_l\left(s_t, z_q\right)-a_t\Vert_2^2 \notag \\
    & + \Vert\operatorname{sg}[z_e]-e\Vert_2^2 + \beta\Vert z_e-\operatorname{sg}[e]\Vert_2^2 \notag \\
    & -\sum_{k=1}^{K} q_\phi(z=k|\boldsymbol{s}_{t,h}, \boldsymbol{a}_{t,h})\log p(k_t=k|s_t),
\end{align}
where $\operatorname{sg}[\cdot]$ is the stopgradient operator, $p(k_t=k|s_t)$ is the probability that $k_t$ is category $k$ and $\beta$ is the hyperparameter of the commitment loss term. Categorical distribution $q_\phi(z=k|\boldsymbol{s}_{t,h}, \boldsymbol{a}_{t,h})$ probabilities are defined as one-hot as
\begin{equation}
    q_\phi(z=k|\mathbf{s}_{t,h},\mathbf{a}_{t,h})=\begin{cases}
        1 & \text{for}\ k=\mathop{\arg\min}\limits_{j}\Vert z_e-e_j\Vert_2, \\
        0 & \text{otherwise,}
    \end{cases}
\end{equation}
where $z_e$ is the output of the encoder.

\subsection{Downstream RL Policy Learning with Skill Prior}
To reuse the learning skill embeddings and improve exploration efficiency in RL training, we utilize the skill prior to regularize the high-level policy in skill space. After pre-training the low-level policy and skill prior, the low-level policy $\pi_l (s_t,z_t)$ and the skill prior $p(k_t|s_t)$ are fixed. The high-level policy $\pi_h (k_t|s_t)$ is executed every $h$-steps, and the output is the categorical distribution of the skill index $k_t$. To obtain the actual skill embedding, we sample an index $k_t$ from the distribution and then query the codebook with $k_t$, i.e., $z_t=Z[k_t]$. In each single time step, the low-level policy predicts the output action conditioned on the state and skill embedding. The low-level policy executes $h$-steps and the state changes to $s_{t+h}$. We use $\mathcal{P}^+ (s_{t+h}|s_t, z_t)$ to denote the state transition after $h$ steps (see line 3-6 in Algorithm \ref{alg}). Next, in order to improve the training efficiency by utilizing the prior information obtained from previous skill training, we add an additional prior term as the learning objective:
\begin{small}
    \begin{equation}\label{eq.rl_objective}
        J(\theta) = \mathbb{E}_{\pi_h} \left[\sum_{t=0}^{H-1} \gamma^t r^+ (s_t,z_t) - \alpha D_{\text{KL}}\left(\pi_h (k_t|s_t)\Vert p(k_t|s_t)\right)\right],
    \end{equation}
\end{small}
where $r^+ (s_t,z_t)=\sum_{i=t}^{t+h}\mathcal{R}(s_i,z_i)$ is the total sum of rewards, $\theta$ is the trainable parameter of policy, $\alpha$ is the hyperparameter of divergence term. To prevent the policy from overfitting the prior, following \cite{pertsch2021accelerating}, $\alpha$ can be automatically tuned by defining a target divergence $\delta$ (see Algorithm \ref{alg} line 15).

Typically, the high-level policy is implemented as a neural network, which is a black box and lacks transparency. In order to achieve an explainable decision making process, we implement the high-level policy $\pi_h (k_t|s_t)$ using a differentiable soft decision tree instead, which can be optimized using existing backpropagation algorithm. Furthermore, the structure of the skill prior is identical to that of the high-level policy, both of which are implemented using soft decision trees. To improve learning efficiency, the parameters of the skill prior are used to initialize the policy during the initialization of training. We modify SAC \cite{haarnoja2018soft} algorithm to learn the objective \ref{eq.rl_objective} and the process is summarized in Algorithm \ref{alg}. See Appendix for more details.

\begin{algorithm}[t]
    \caption{SkillTree RL Training}
    \label{alg}
    \textbf{Inputs}: codebook $Z$, target divergence $\delta$, learning rates $\lambda_\theta, \lambda_\psi, \lambda_\alpha$, target update rate $\tau$, DT skill prior $p(k_t|s_t)$ and discount factor $\gamma$.
    \begin{algorithmic}[1] 
        \STATE Initialize replay buffer $\mathcal{B}$, $d$-depth high-level DT policy $\pi_\theta(k_t|s_t)$, critic $Q_\psi (s_t,z_t)$, target network $Q_{\bar{\psi}}(s_t,z_t)$
        \FOR {each iteration}
            \FOR {every $h$ environment steps}
                \STATE $k_t\sim \pi_\theta(k_t|s_t)$
                \STATE $z_t=Z[k_t]$
                \STATE $s_{t'}\sim \mathcal{P}^+ (s_{t+h}|s_t,z_t)$
                \STATE $\mathcal{B}\leftarrow \mathcal{B}\cup \{s_t,z_t,r_t^+ ,s_{t'}\}$
            \ENDFOR
            \FOR {each gradient step}
                \STATE $k_{t'}\sim \pi_\theta(k_{t'}|s_{t'})$
                \STATE $z_{t'}=Z[{k_t'}]$
                \STATE \begin{small}$\bar{Q}=r_t^+ +\gamma\left[Q_{\bar{\psi}}(s_{t'},z_{t'})+\alpha D_{\text{KL}}(\pi_\theta (k_{t'}|s_{t'})\Vert p(k_{t'}|s_{t'}))\right]$\end{small}
                \STATE \begin{small}$\theta\leftarrow \theta-\lambda_\theta\nabla_\theta \left[Q_{\bar{\psi}}(s_t,z_t)-D_{\text{KL}}(\pi_\theta (k_t|s_t)\Vert p(k_t|s_t))\right]$\end{small}
                \STATE $\psi\leftarrow \psi-\lambda_\psi \nabla_\psi\left[\frac{1}{2}\left(Q_\psi (s_t|z_t)-\bar{Q}\right)^2\right]$
                \STATE $\alpha\leftarrow \alpha-\lambda_\alpha \nabla_\alpha\left[\alpha\left(D_{\text{KL}}(\pi_\theta (k_t|s_t)\Vert p(k_t|s_t))-\delta\right)\right]$
                \STATE $\bar{\psi}\leftarrow \tau\psi+(1-\tau)\bar{\psi}$
            \ENDFOR
        \ENDFOR
        \STATE \textbf{return} policy $\pi_\theta (k_t|s_t)$, finetuned codebook $Z$
    \end{algorithmic}
\end{algorithm}

\subsection{Distilling the Decision Tree}
After downstream RL training, we obtain a soft decision tree with skill-level explainability, where each decision node probabilistically selects child nodes until a skill choice is made. Generally, models with fewer parameters are easier to understand and accept. To further simplify the model structure, we distill the soft decision tree into a more straightforward hard decision tree through discretization. Previous methods select the feature with the highest weight at each node but it often obviously degrades tree performance \cite{ding2020cdt}. By framing the problem as an imitation learning task, we leverage the learned high-level policy as an expert policy. We sample trajectories from the environment to generate the dataset. Given the simplified skill space, we employ a low-depth CART to classify state-skill index pairs. This approach effectively manages the complexity of the tree while preserving performance, ensuring that the model remains both explainable and efficient.

\section{Experiments}\label{sec.experiments}
In this section, we evaluate SkillTree’s effectiveness in achieving a balance between explainability and performance. We demonstrate that our method not only provides clear, skill-level explanations but also achieves performance on par with neural network based approaches.

\subsection{Tasks Setup}
\begin{figure}[t]
    \centering
    \subfigure[Kitchen MKBL]{
        \begin{minipage}[t]{0.47\linewidth}
            \centering\includegraphics[width=\linewidth]{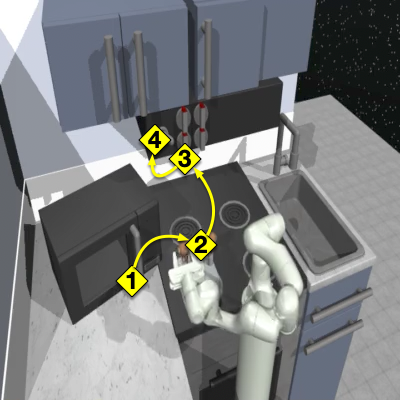}
        \end{minipage}\label{fig.mkbl}
    }
    \subfigure[Kitchen MLSH]{
        \begin{minipage}[t]{0.47\linewidth}
            \centering\includegraphics[width=\linewidth]{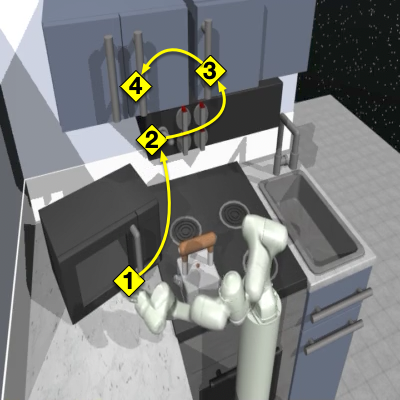}
        \end{minipage}\label{fig.mlsh}
    }
    \subfigure[CALVIN]{
        \begin{minipage}[t]{0.47\linewidth}
            \centering\includegraphics[width=\linewidth]{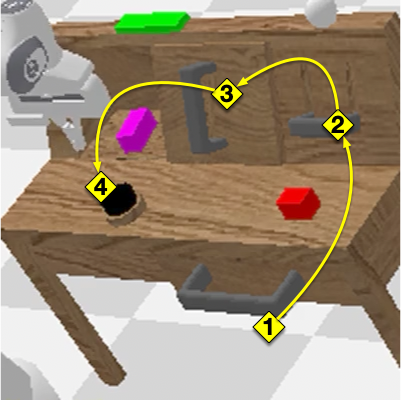}
        \end{minipage}
    }
    \subfigure[Office]{
        \begin{minipage}[t]{0.47\linewidth}
            \centering\includegraphics[width=\linewidth]{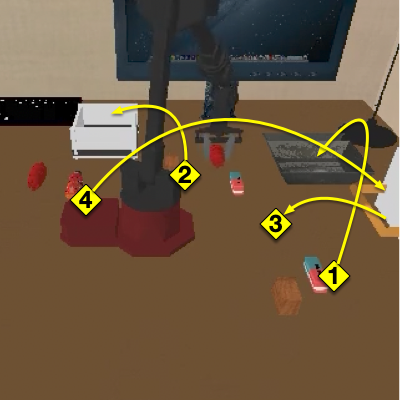}
        \end{minipage}
    }
    \caption{Four long-horizon sparse reward tasks to evaluate. (a) The robotic arm has to finish four subtasks in the correct order, i.e., \textit{Microwave - Kettle - Bottom Burner - Light} (\textit{MKBL}). (b) Similar to (a), but with different subtasks: \textit{Microwave - Ligt - Slide Cabinet - Hinge Cabinet} (\textit{MLSH}). (c) Finish subtasks in the correct order, i.e., \textit{Open Drawer - Turn on Lightbulb - Move Slider Left - Turn on LED}. (d) In the office cleaning task, the robotic arm needs to pick up objects and place them in corresponding containers in sequence.}
    \label{fig.environments}
\end{figure}

To evaluate the performance of our method in long-horizon sparse reward tasks, we chose the Franka Kitchen control tasks in D4RL \cite{fu2020d4rl}, robotic arm manipulation task in CALVIN \cite{mees2022calvin} and office cleaning task \cite{pertsch2021demonstration}, as illustrated in Figure \ref{fig.environments}.

\begin{figure*}[t]
    \centering
    \includegraphics[width=\linewidth]{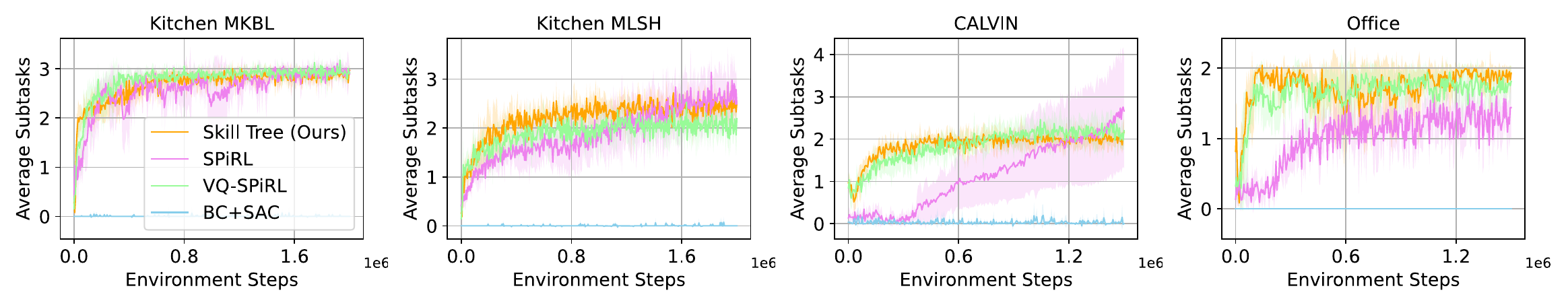}
    \caption{Downstream task learning curves of both our method and baselines. Averaged over 5 independent runs.}
    \label{fig.curve}
\end{figure*}

\begin{table*}
    \caption{Average completed subtasks (ACS) and the number of leaf nodes (Leaf) across methods and domains.}
    \centering
    \begin{tabular}{c|cc|cc|cc|cc}
    \toprule
    \multicolumn{1}{c}{\multirow{2}{*}{Method}} \vline & \multicolumn{2}{c}{Kitchen MKBL} \vline & \multicolumn{2}{c}{Kitchen MLSH} \vline & \multicolumn{2}{c}{CALVIN} \vline & \multicolumn{2}{c}{Office} \\ \cline{2-9}
    \multicolumn{1}{c}{} \vline & ACS & Leaf & ACS & Leaf & ACS & Leaf & ACS & Leaf \\
    \midrule
    SPiRL & 3.08 $\pm$ 0.46 & N/A & \textbf{2.82} $\pm$ \textbf{1.08} & N/A & 2.77 $\pm$ 1.46 & N/A & 1.48 $\pm$ 0.96 & N/A \\
    VQ-SPiRL & 2.97 $\pm$ 0.41 & N/A & 2.04 $\pm$ 0.85 & N/A & 2.10 $\pm$ 0.81 & N/A & 1.69 $\pm$ 0.64 & N/A \\
    BC+SAC & 0.00 $\pm$ 0.00 & N/A & 0.00 $\pm$ 0.00 & N/A & 0.01
     $\pm$ 0.10 & N/A & 0.00 $\pm$ 0.00 & N/A \\
    CART & 2.11 $\pm$ 0.92 & 987 & 1.16 $\pm$ 0.78 & 981 & 1.02 $\pm$ 0.20 & 998 & 0.00 $\pm$ 0.00 & 949 \\
    \midrule
    SkillTree & 2.90 $\pm$ 0.42 & 64 & 2.58 $\pm$ 0.80 & 64 & 1.90 $\pm$ 0.90 & 64 & \textbf{2.00} $\pm$ \textbf{0.00} & 64 \\
    SkillTree (Distilling) & 3.06 $\pm$ 0.65 & 64 & 2.52 $\pm$ 0.82 & 64 & 2.42 $\pm$ 0.49 & 64 & \textbf{2.00} $\pm$ \textbf{0.00} & 64 \\
    SkillTree (DC + Distilling) & \textbf{3.25} $\pm$ \textbf{0.70} & 64 & 2.62 $\pm$ 0.76 & 64 & \textbf{3.00} $\pm$ \textbf{0.00} & 64 & \textbf{2.00} $\pm$ \textbf{0.00} & 64 \\
    \bottomrule
    \end{tabular}
    \label{tbl.performance}
\end{table*}

\subsubsection{Kitchen}
The Franka Kitchen environment is a robotic arm controlled environment based on the mujoco implementation. The environment contains a 7-DOF robotic arm that is able to interact with other objects in the kitchen, such as a kettle and other objects. A total of 7 subtasks are included. The observation is 60-dimensional and contains information about joint positions as well as task parameters, and the action space is 9-dimensional with episode length of 280. The offset dataset contains 601 trajectories, and each trajectory completes a variety of tasks that may not be in the same order, but no more than 3 subtasks. We set two sets of different subtasks: \textit{MKBL} and \textit{MLSH}, as shown in Figure \ref{fig.mkbl} and \ref{fig.mlsh}. The \textit{MLSH} task is more difficult to learn due to the very low frequency of subtask transitions in the dataset.

\subsubsection{CALVIN}
In CALVIN, a 7-DOF Franka Emika Panda robotic arm needs to be controlled to open a drawer, light a light bulb, move a slider to the left and light an LED in sequence. the observation space is 21-dimensional and contains both robotic arm states and object states with episode length of 360 steps. The dataset contains 1,239 trajectories and a total of 34 subtasks. Since it contains a large number of subtasks, the average frequency of transitions between any subtasks in the dataset is very low, requiring more precise skill selection, which poses a greater challenge during the learning of the high-level policy.

\subsubsection{Office}
For the Office environment, a 5-DOF WidowX robotic arm needs to be controlled to pick up multiple objects and place them in their corresponding containers with episode length of 350. We utilized a dataset collected by \cite{pertsch2021demonstration}, which includes 2,331 trajectories. The state space is 97-dimensional and the action space is 8-dimensional. Due to the freedom of object manipulation in this environment, the task is more challenging to complete.

\subsection{Baselines}
In the experiments, we compare our proposed method against several representative baselines to demonstrate its effectiveness and competitive performance with additional skill-level explainability.

\begin{itemize}
    \item SPiRL \cite{pertsch2021accelerating}: Learns continuous skills embeddings and a skill prior, and guides a high-level policy using the learned prior. It serves as a competitive skill-based RL algorithm but lacks explainability.
    \item VQ-SPiRL: Discretizes the skills in SPiRL but uses neural networks as the high-level policy model structure.
    \item Behavioral Cloning + Soft Actor-Critic (BC+SAC): Trains a supervised behavioral cloning policy from the demonstration dataset and finetunes it on the downstream task using Soft Actor-Critic \cite{haarnoja2018soft}. It serves as a general algorithm without leveraging skills.
    \item CART \cite{loh2011classification}: Imitation learning using CART with trained SPiRL agent as the teacher.
\end{itemize}

For SkillTree, we set the depth of the tree to 6 in all domains. The size of codebook $K$ is 16 for \textit{Kitchen} and \textit{Office}, and 8 for \textit{CALVIN}. For CART, we sample 1,000 trajectories on each domain to train and set the maximum depth to 10.

\subsection{Results}\label{ssec.results}
From the learning curves shown in Figure \ref{fig.curve}, it is evident that our method demonstrates comparable learning efficiency and asymptotic performance in all domains. This highlights the effectiveness of our discrete skill representation in handling complex, long-horizon tasks. BC+SAC failed to complete any subtasks because of the sparse reward. In the \textit{CALVIN} and \textit{Office} domain, our method exhibits a marked improvement in early learning efficiency compared to SPiRL and performs comparably to VQ-SPiRL. This advantage underscores the benefit of discrete skills in enhancing exploration efficiency, allowing our approach to quickly identify and leverage useful skills for target task. Through comparing to VQ-SPiRL, we validate that the decision tree can maintain performance while reducing model complexity in simplified discrete skill space.


In Table \ref{tbl.performance}, we evaluate the additional SkillTree after distilling on sampled dataset. Specifically, we use two sampling methods. One is to sample the trajectories directly with SkillTree (distilling, D), and the other is to clean the trajectories and keep only those with higher rewards (data cleaning + distilling, DC+D). In data cleaning, we select the trajectories completed subtasks greater than 2 for \textit{Kitchen} and \textit{CALVIN}, and greater than 1 for \textit{Office}. In each environment, we sample 1000 trajectories. To control the complexity, we set the depth of the tree to 6 in all domains. We observe that after data cleaning and distillation, the performance of SkillTree can be further enhanced. In most of the domains, SkillTree DC+D achieves best performance than other baselines. This improvement reflects the effectiveness of our method in optimizing the skill representations, allowing it to adapt well even in complex scenarios. Moreover, the distillation process with data cleaning not only simplifies the model but also ensures that it retains critical decision making capabilities, leading to better performance across different tasks. In contrast, CART has worse performance despite having deeper layers and a much larger number of leaf nodes than SkillTree. Huge number of leaf nodes ($\sim$ 1000) suggests that it is difficult to understand intuitively.

\subsection{Explainability}
\begin{figure}[t]
    \centering
    \includegraphics[width=\linewidth]{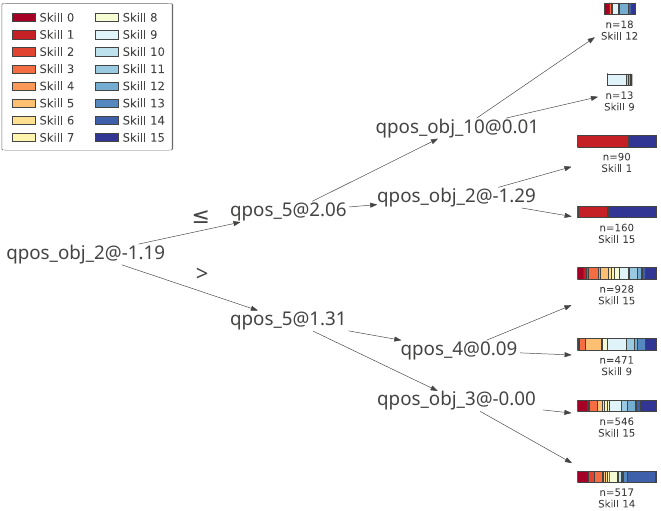}
    \caption{Visualization of the SkillTree (DC+D) with depth 3. \texttt{qpos} and \texttt{qpos\_obj} mean the position of the robotic arm and objects, respectively. $n$ denotes the number of state-skill pairs in the divided decision set.}
    \label{fig.hdt}
\end{figure}

\begin{figure}[t] 
    \centering
    \includegraphics[width=\linewidth]{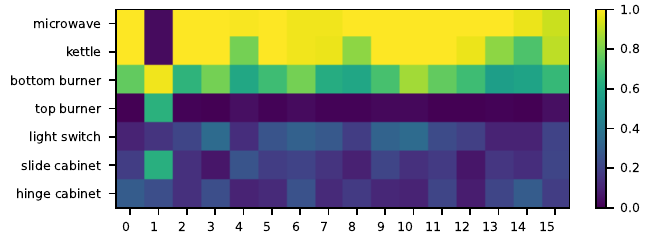}
    \caption{We fix the skill output of high-level policy and evaluate the focus on subtasks of each skill in \textit{Kitchen MKBL} task. We set the number of skills $K$ to 16, as shown in the x-axis. The color means the success rate of each subtask.}
    \label{fig.skill_eval}
\end{figure}

\begin{figure}[t]
    \centering
    \includegraphics[width=\linewidth]{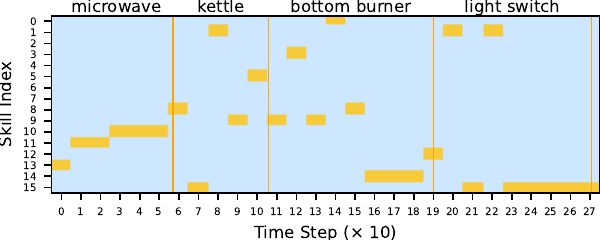}
    \caption{Skill index output visualization of an episode in \textit{Kitchen MKBL} task, which finished all 4 subtasks.}
    \label{fig.episode}
\end{figure}

In Figure \ref{fig.hdt}, we visualize the final decision tree obtained after distillation for the \textit{Kitchen MKBL} task. This depth of decision tree is only three but can complete an average of 3.03 subtasks. The very few parameters make the decision-making process clear and easy to understand. Each node makes splits based on one of the observed features. In the 60-dimensional observations of the \textit{Kitchen} domain, the last 30 dimensions are task parameters that are independent of the state and thus do not influence decision making. This property is reflected in the learned tree structure because the decision nodes use only the features in the first 30 dimensions (robot arm position and object position) for splitting.

To analyze the effects of each skill, we also evaluated the effectiveness of different skills, as shown in Figure \ref{fig.skill_eval}. We fixed the skill output individually and evaluated the completion of subtasks. Each skill is capable of completing different subtasks with varying focus. SkillTree selects the appropriate skill based on the observed state, which are then executed by the low-level policy. Figure \ref{fig.episode} illustrates the sequence of skill indices chosen in one episode, during which four subtasks were completed. It can be seen that the same skill was repeatedly selected at consecutive time steps to complete specific subtasks, indicating the reuse of particular skills. Moreover, repeated selection of the same skill indicates a relatively long-term and stable skill execution process. By isolating and testing each skill, we can clearly see which skills are most effective for particular subtasks. This makes it easier to understand the role and functionality of each skill within the decision making process. See Appendix for more experiment results as well as visualizations.

\section{Conclusions and Limitations}
We proposed a novel hierarchical skill-based explainable reinforcement learning framework designed to tackle the challenges of long-hoziron decision making in continuous control tasks. By integrating discrete skill representations and DT, our method offers a robust solution that enhances explainability without sacrificing performance. The proposed approach effectively regularizes the skill space, simplifies policy learning, and leverages the transparency of the DT. Experimental results demonstrate that our method not only achieves competitive performance compared to neural network based approaches but also provides valuable skill-level explanations. This work represents a important step towards more transparent and explainable RL, paving the way for their broader application in complex real-world scenarios where understanding the decision making process is essential. Despite the promising results demonstrated, several limitations remain. The low-level policy in SkillTree still lacks explainability due to its reliance on neural networks, which limits full transparency in the decision-making process. Additionally, the current approach assumes an offline, task-agnostic dataset for skill learning, which may not always be available in practical scenarios. Future work could focus on developing methods to enhance the explainability of low-level policies, possibly by integrating explainable models directly into the low-level decision-making process.


\bibliography{aaai25}

\begin{thebibliography}{54}
\providecommand{\natexlab}[1]{#1}

\bibitem[{Akkaya et~al.(2019)Akkaya, Andrychowicz, Chociej, Litwin, McGrew, Petron, Paino, Plappert, Powell, Ribas et~al.}]{akkaya2019solving}
Akkaya, I.; Andrychowicz, M.; Chociej, M.; Litwin, M.; McGrew, B.; Petron, A.; Paino, A.; Plappert, M.; Powell, G.; Ribas, R.; et~al. 2019.
\newblock Solving rubik's cube with a robot hand.
\newblock \emph{arXiv preprint arXiv:1910.07113}.

\bibitem[{Anderson et~al.(2019)Anderson, Dodge, Sadarangani, Juozapaitis, Newman, Irvine, Chattopadhyay, Fern, and Burnett}]{anderson2019explaining}
Anderson, A.; Dodge, J.; Sadarangani, A.; Juozapaitis, Z.; Newman, E.; Irvine, J.; Chattopadhyay, S.; Fern, A.; and Burnett, M. 2019.
\newblock Explaining reinforcement learning to mere mortals: an empirical study.
\newblock In \emph{Proceedings of the 28th International Joint Conference on Artificial Intelligence}, 1328--1334.

\bibitem[{BAAI(2023)}]{baai2023plan4mc}
BAAI, P. 2023.
\newblock Plan4mc: Skill reinforcement learning and planning for open-world minecraft tasks.
\newblock \emph{arXiv preprint arXiv:2303.16563}.

\bibitem[{Bastani, Pu, and Solar-Lezama(2018)}]{bastani2018verifiable}
Bastani, O.; Pu, Y.; and Solar-Lezama, A. 2018.
\newblock Verifiable reinforcement learning via policy extraction.
\newblock In \emph{Advances in Neural Information Processing Systems}, volume~31.

\bibitem[{Bewley and Lawry(2021)}]{bewley2021tripletree}
Bewley, T.; and Lawry, J. 2021.
\newblock Tripletree: A versatile interpretable representation of black box agents and their environments.
\newblock In \emph{Proceedings of the AAAI Conference on Artificial Intelligence}, volume~35, 11415--11422.

\bibitem[{Cheng et~al.(2024)Cheng, Wu, Yu, Sun, Guo, and Xing}]{cheng2024statemask}
Cheng, Z.; Wu, X.; Yu, J.; Sun, W.; Guo, W.; and Xing, X. 2024.
\newblock Statemask: Explaining deep reinforcement learning through state mask.
\newblock In \emph{Advances in Neural Information Processing Systems}, volume~36.

\bibitem[{Coppens et~al.(2019)Coppens, Efthymiadis, Lenaerts, Now{\'e}, Miller, Weber, and Magazzeni}]{coppens2019distilling}
Coppens, Y.; Efthymiadis, K.; Lenaerts, T.; Now{\'e}, A.; Miller, T.; Weber, R.; and Magazzeni, D. 2019.
\newblock Distilling deep reinforcement learning policies in soft decision trees.
\newblock In \emph{Proceedings of the IJCAI 2019 workshop on explainable artificial intelligence}, 1--6.

\bibitem[{Costa and Pedreira(2023)}]{costa2023recent}
Costa, V.~G.; and Pedreira, C.~E. 2023.
\newblock Recent advances in decision trees: An updated survey.
\newblock \emph{Artificial Intelligence Review}, 56(5): 4765--4800.

\bibitem[{Dalal, Pathak, and Salakhutdinov(2021)}]{dalal2021accelerating}
Dalal, M.; Pathak, D.; and Salakhutdinov, R.~R. 2021.
\newblock Accelerating robotic reinforcement learning via parameterized action primitives.
\newblock In \emph{Advances in Neural Information Processing Systems}, volume~34, 21847--21859.

\bibitem[{Deshmukh et~al.(2023)Deshmukh, Dasgupta, Krishnamurthy, Jiang, Agarwal, Theocharous, and Subramanian}]{deshmukh2023explaining}
Deshmukh, S.~V.; Dasgupta, A.; Krishnamurthy, B.; Jiang, N.; Agarwal, C.; Theocharous, G.; and Subramanian, J. 2023.
\newblock Explaining RL Decisions with Trajectories.
\newblock In \emph{International Conference on Learning Representations}.

\bibitem[{Dhebar and Deb(2020)}]{dhebar2020interpretable}
Dhebar, Y.; and Deb, K. 2020.
\newblock Interpretable rule discovery through bilevel optimization of split-rules of nonlinear decision trees for classification problems.
\newblock \emph{IEEE Transactions on Cybernetics}, 51(11): 5573--5584.

\bibitem[{Ding et~al.(2020)Ding, Hernandez-Leal, Ding, Li, and Huang}]{ding2020cdt}
Ding, Z.; Hernandez-Leal, P.; Ding, G.~W.; Li, C.; and Huang, R. 2020.
\newblock Cdt: Cascading decision trees for explainable reinforcement learning.
\newblock \emph{arXiv preprint arXiv:2011.07553}.

\bibitem[{Frosst and Hinton(2017)}]{frosst2017distilling}
Frosst, N.; and Hinton, G. 2017.
\newblock Distilling a neural network into a soft decision tree.
\newblock \emph{arXiv preprint arXiv:1711.09784}.

\bibitem[{Fu et~al.(2020)Fu, Kumar, Nachum, Tucker, and Levine}]{fu2020d4rl}
Fu, J.; Kumar, A.; Nachum, O.; Tucker, G.; and Levine, S. 2020.
\newblock D4rl: Datasets for deep data-driven reinforcement learning.
\newblock \emph{arXiv preprint arXiv:2004.07219}.

\bibitem[{Greydanus et~al.(2018)Greydanus, Koul, Dodge, and Fern}]{greydanus2018visualizing}
Greydanus, S.; Koul, A.; Dodge, J.; and Fern, A. 2018.
\newblock Visualizing and understanding atari agents.
\newblock In \emph{International conference on machine learning}, 1792--1801. PMLR.

\bibitem[{Guo et~al.(2021)Guo, Wu, Khan, and Xing}]{guo2021edge}
Guo, W.; Wu, X.; Khan, U.; and Xing, X. 2021.
\newblock Edge: Explaining deep reinforcement learning policies.
\newblock In \emph{Advances in Neural Information Processing Systems}, volume~34, 12222--12236.

\bibitem[{Haarnoja et~al.(2018)Haarnoja, Zhou, Abbeel, and Levine}]{haarnoja2018soft}
Haarnoja, T.; Zhou, A.; Abbeel, P.; and Levine, S. 2018.
\newblock Soft Actor-Critic: Off-Policy Maximum Entropy Deep Reinforcement Learning with a Stochastic Actor.
\newblock In Dy, J.; and Krause, A., eds., \emph{Proceedings of the 35th International Conference on Machine Learning}, volume~80 of \emph{Proceedings of Machine Learning Research}, 1861--1870. PMLR.

\bibitem[{Hausman et~al.(2018)Hausman, Springenberg, Wang, Heess, and Riedmiller}]{hausman2018learning}
Hausman, K.; Springenberg, J.~T.; Wang, Z.; Heess, N.; and Riedmiller, M. 2018.
\newblock Learning an embedding space for transferable robot skills.
\newblock In \emph{International Conference on Learning Representations}.

\bibitem[{Hein, Udluft, and Runkler(2018)}]{hein2018interpretable}
Hein, D.; Udluft, S.; and Runkler, T.~A. 2018.
\newblock Interpretable policies for reinforcement learning by genetic programming.
\newblock \emph{Engineering Applications of Artificial Intelligence}, 76: 158--169.

\bibitem[{Heuillet, Couthouis, and D{\'\i}az-Rodr{\'\i}guez(2021)}]{heuillet2021explainability}
Heuillet, A.; Couthouis, F.; and D{\'\i}az-Rodr{\'\i}guez, N. 2021.
\newblock Explainability in deep reinforcement learning.
\newblock \emph{Knowledge-Based Systems}, 214: 106685.

\bibitem[{Heuillet, Couthouis, and D{\'\i}az-Rodr{\'\i}guez(2022)}]{heuillet2022collective}
Heuillet, A.; Couthouis, F.; and D{\'\i}az-Rodr{\'\i}guez, N. 2022.
\newblock Collective explainable AI: Explaining cooperative strategies and agent contribution in multiagent reinforcement learning with shapley values.
\newblock \emph{IEEE Computational Intelligence Magazine}, 17(1): 59--71.

\bibitem[{Hickling et~al.(2023)Hickling, Zenati, Aouf, and Spencer}]{hickling2023explainability}
Hickling, T.; Zenati, A.; Aouf, N.; and Spencer, P. 2023.
\newblock Explainability in deep reinforcement learning: A review into current methods and applications.
\newblock \emph{ACM Computing Surveys}, 56(5): 1--35.

\bibitem[{Jitosho et~al.(2023)Jitosho, Lum, Okamura, and Liu}]{jitosho2023reinforcement}
Jitosho, R.; Lum, T. G.~W.; Okamura, A.; and Liu, K. 2023.
\newblock Reinforcement Learning Enables Real-Time Planning and Control of Agile Maneuvers for Soft Robot Arms.
\newblock In Tan, J.; Toussaint, M.; and Darvish, K., eds., \emph{Proceedings of The 7th Conference on Robot Learning}, volume 229 of \emph{Proceedings of Machine Learning Research}, 1131--1153. PMLR.

\bibitem[{Kipf et~al.(2019)Kipf, Li, Dai, Zambaldi, Sanchez-Gonzalez, Grefenstette, Kohli, and Battaglia}]{kipf2019compile}
Kipf, T.; Li, Y.; Dai, H.; Zambaldi, V.; Sanchez-Gonzalez, A.; Grefenstette, E.; Kohli, P.; and Battaglia, P. 2019.
\newblock Compile: Compositional imitation learning and execution.
\newblock In \emph{International Conference on Machine Learning}, 3418--3428. PMLR.

\bibitem[{Kulh{\'a}nek, Derner, and Babu{\v{s}}ka(2021)}]{kulhanek2021visual}
Kulh{\'a}nek, J.; Derner, E.; and Babu{\v{s}}ka, R. 2021.
\newblock Visual navigation in real-world indoor environments using end-to-end deep reinforcement learning.
\newblock \emph{IEEE Robotics and Automation Letters}, 6(3): 4345--4352.

\bibitem[{Landajuela et~al.(2021)Landajuela, Petersen, Kim, Santiago, Glatt, Mundhenk, Pettit, and Faissol}]{landajuela2021discovering}
Landajuela, M.; Petersen, B.~K.; Kim, S.; Santiago, C.~P.; Glatt, R.; Mundhenk, N.; Pettit, J.~F.; and Faissol, D. 2021.
\newblock Discovering symbolic policies with deep reinforcement learning.
\newblock In Meila, M.; and Zhang, T., eds., \emph{Proceedings of the 38th International Conference on Machine Learning}, volume 139 of \emph{Proceedings of Machine Learning Research}, 5979--5989. PMLR.

\bibitem[{Lee, Yang, and Lim(2019)}]{lee2019learning}
Lee, Y.; Yang, J.; and Lim, J.~J. 2019.
\newblock Learning to coordinate manipulation skills via skill behavior diversification.
\newblock In \emph{International conference on learning representations}.

\bibitem[{Liu et~al.(2019)Liu, Schulte, Zhu, and Li}]{liu2019toward}
Liu, G.; Schulte, O.; Zhu, W.; and Li, Q. 2019.
\newblock Toward interpretable deep reinforcement learning with linear model u-trees.
\newblock In \emph{Machine Learning and Knowledge Discovery in Databases: European Conference, ECML PKDD 2018, Dublin, Ireland, September 10--14, 2018, Proceedings, Part II 18}, 414--429. Springer.

\bibitem[{Liu et~al.(2021)Liu, Sun, Schulte, and Poupart}]{liu2021learning}
Liu, G.; Sun, X.; Schulte, O.; and Poupart, P. 2021.
\newblock Learning tree interpretation from object representation for deep reinforcement learning.
\newblock In \emph{Advances in Neural Information Processing Systems}, volume~34, 19622--19636.

\bibitem[{Loh(2011)}]{loh2011classification}
Loh, W.-Y. 2011.
\newblock Classification and regression trees.
\newblock \emph{Wiley interdisciplinary reviews: data mining and knowledge discovery}, 1(1): 14--23.

\bibitem[{Lynch et~al.(2020)Lynch, Khansari, Xiao, Kumar, Tompson, Levine, and Sermanet}]{lynch2020learning}
Lynch, C.; Khansari, M.; Xiao, T.; Kumar, V.; Tompson, J.; Levine, S.; and Sermanet, P. 2020.
\newblock Learning latent plans from play.
\newblock In \emph{Conference on robot learning}, 1113--1132. PMLR.

\bibitem[{Lyu et~al.(2019)Lyu, Yang, Liu, and Gustafson}]{lyu2019sdrl}
Lyu, D.; Yang, F.; Liu, B.; and Gustafson, S. 2019.
\newblock SDRL: interpretable and data-efficient deep reinforcement learning leveraging symbolic planning.
\newblock In \emph{Proceedings of the AAAI Conference on Artificial Intelligence}, volume~33, 2970--2977.

\bibitem[{Ma et~al.(2021)Ma, Zhuang, Weng, Zhuo, Li, Liu, and Hao}]{ma2021learning}
Ma, Z.; Zhuang, Y.; Weng, P.; Zhuo, H.~H.; Li, D.; Liu, W.; and Hao, J. 2021.
\newblock Learning symbolic rules for interpretable deep reinforcement learning.
\newblock \emph{arXiv preprint arXiv:2103.08228}.

\bibitem[{Madumal et~al.(2020)Madumal, Miller, Sonenberg, and Vetere}]{madumal2020explainable}
Madumal, P.; Miller, T.; Sonenberg, L.; and Vetere, F. 2020.
\newblock Explainable reinforcement learning through a causal lens.
\newblock In \emph{Proceedings of the AAAI conference on artificial intelligence}, volume~34, 2493--2500.

\bibitem[{Mees et~al.(2022)Mees, Hermann, Rosete-Beas, and Burgard}]{mees2022calvin}
Mees, O.; Hermann, L.; Rosete-Beas, E.; and Burgard, W. 2022.
\newblock Calvin: A benchmark for language-conditioned policy learning for long-horizon robot manipulation tasks.
\newblock \emph{IEEE Robotics and Automation Letters}, 7(3): 7327--7334.

\bibitem[{Milani et~al.(2024)Milani, Topin, Veloso, and Fang}]{milani2024explainable}
Milani, S.; Topin, N.; Veloso, M.; and Fang, F. 2024.
\newblock Explainable reinforcement learning: A survey and comparative review.
\newblock \emph{ACM Computing Surveys}, 56(7): 1--36.

\bibitem[{Mnih et~al.(2015)Mnih, Kavukcuoglu, Silver, Rusu, Veness, Bellemare, Graves, Riedmiller, Fidjeland, Ostrovski et~al.}]{mnih2015human}
Mnih, V.; Kavukcuoglu, K.; Silver, D.; Rusu, A.~A.; Veness, J.; Bellemare, M.~G.; Graves, A.; Riedmiller, M.; Fidjeland, A.~K.; Ostrovski, G.; et~al. 2015.
\newblock Human-level control through deep reinforcement learning.
\newblock \emph{nature}, 518(7540): 529--533.

\bibitem[{Molnar, Casalicchio, and Bischl(2020)}]{molnar2020interpretable}
Molnar, C.; Casalicchio, G.; and Bischl, B. 2020.
\newblock Interpretable machine learning--a brief history, state-of-the-art and challenges.
\newblock In \emph{Joint European conference on machine learning and knowledge discovery in databases}, 417--431. Springer.

\bibitem[{Olson et~al.(2021)Olson, Khanna, Neal, Li, and Wong}]{olson2021counterfactual}
Olson, M.~L.; Khanna, R.; Neal, L.; Li, F.; and Wong, W.-K. 2021.
\newblock Counterfactual state explanations for reinforcement learning agents via generative deep learning.
\newblock \emph{Artificial Intelligence}, 295: 103455.

\bibitem[{Orfanos and Lelis(2023)}]{orfanos2023synthesizing}
Orfanos, S.; and Lelis, L.~H. 2023.
\newblock Synthesizing programmatic policies with actor-critic algorithms and relu networks.
\newblock \emph{arXiv preprint arXiv:2308.02729}.

\bibitem[{Pertsch, Lee, and Lim(2021)}]{pertsch2021accelerating}
Pertsch, K.; Lee, Y.; and Lim, J. 2021.
\newblock Accelerating reinforcement learning with learned skill priors.
\newblock In \emph{Conference on robot learning}, 188--204. PMLR.

\bibitem[{Pertsch et~al.(2021)Pertsch, Lee, Wu, and Lim}]{pertsch2021demonstration}
Pertsch, K.; Lee, Y.; Wu, Y.; and Lim, J.~J. 2021.
\newblock Demonstration-Guided Reinforcement Learning with Learned Skills.
\newblock In \emph{5th Annual Conference on Robot Learning}.

\bibitem[{Quinlan(1993)}]{quinlan1993c4}
Quinlan, J.~R. 1993.
\newblock C4. 5: Programs for Machine Learning.

\bibitem[{Rajeswaran et~al.(2017)Rajeswaran, Lowrey, Todorov, and Kakade}]{rajeswaran2017towards}
Rajeswaran, A.; Lowrey, K.; Todorov, E.~V.; and Kakade, S.~M. 2017.
\newblock Towards generalization and simplicity in continuous control.
\newblock In \emph{Advances in neural information processing systems}, volume~30.

\bibitem[{Shankar and Gupta(2020)}]{shankar2020learning}
Shankar, T.; and Gupta, A. 2020.
\newblock Learning robot skills with temporal variational inference.
\newblock In \emph{International Conference on Machine Learning}, 8624--8633. PMLR.

\bibitem[{Shi, Lim, and Lee(2023)}]{shi2023skill}
Shi, L.~X.; Lim, J.~J.; and Lee, Y. 2023.
\newblock Skill-based Model-based Reinforcement Learning.
\newblock In \emph{Conference on Robot Learning}, 2262--2272. PMLR.

\bibitem[{Shiarlis et~al.(2018)Shiarlis, Wulfmeier, Salter, Whiteson, and Posner}]{shiarlis2018taco}
Shiarlis, K.; Wulfmeier, M.; Salter, S.; Whiteson, S.; and Posner, I. 2018.
\newblock Taco: Learning task decomposition via temporal alignment for control.
\newblock In \emph{International Conference on Machine Learning}, 4654--4663. PMLR.

\bibitem[{Shu, Xiong, and Socher(2018)}]{shu2018hierarchical}
Shu, T.; Xiong, C.; and Socher, R. 2018.
\newblock Hierarchical and Interpretable Skill Acquisition in Multi-task Reinforcement Learning.
\newblock In \emph{International Conference on Learning Representations}.

\bibitem[{Silva et~al.(2020)Silva, Gombolay, Killian, Jimenez, and Son}]{silva2020optimization}
Silva, A.; Gombolay, M.; Killian, T.; Jimenez, I.; and Son, S.-H. 2020.
\newblock Optimization methods for interpretable differentiable decision trees applied to reinforcement learning.
\newblock In \emph{International conference on artificial intelligence and statistics}, 1855--1865. PMLR.

\bibitem[{Sutton and Barto(2018)}]{sutton2018reinforcement}
Sutton, R.~S.; and Barto, A.~G. 2018.
\newblock \emph{Reinforcement Learning: An Introduction}.
\newblock MIT press.

\bibitem[{Van Den~Oord, Vinyals et~al.(2017)}]{van2017neural}
Van Den~Oord, A.; Vinyals, O.; et~al. 2017.
\newblock Neural discrete representation learning.
\newblock In \emph{Advances in Neural Information Processing Systems}, volume~30.

\bibitem[{Vasić et~al.(2022)Vasić, Petrović, Wang, Nikolić, Singh, and Khurshid}]{vasicETAL22MoET}
Vasić, M.; Petrović, A.; Wang, K.; Nikolić, M.; Singh, R.; and Khurshid, S. 2022.
\newblock MoËT: Mixture of Expert Trees and its application to verifiable reinforcement learning.
\newblock \emph{Neural Networks}, 151: 34--47.

\bibitem[{Wabartha and Pineau(2023)}]{wabartha2023piecewise}
Wabartha, M.; and Pineau, J. 2023.
\newblock Piecewise Linear Parametrization of Policies: Towards Interpretable Deep Reinforcement Learning.
\newblock In \emph{International Conference on Learning Representations}.

\bibitem[{Zhang et~al.(2021)Zhang, Zhang, Xu, Gao, and Gao}]{zhang2021explainable}
Zhang, K.; Zhang, J.; Xu, P.-D.; Gao, T.; and Gao, D.~W. 2021.
\newblock Explainable AI in deep reinforcement learning models for power system emergency control.
\newblock \emph{IEEE Transactions on Computational Social Systems}, 9(2): 419--427.

\end{thebibliography}

\end{document}